\documentclass[10pt,twocolumn,letterpaper]{article}

\usepackage{iccv}
\usepackage{times}
\usepackage{epsfig}
\usepackage{graphicx}
\usepackage{amsmath}
\usepackage{amssymb}

\usepackage{array}
\usepackage{algorithm}
\usepackage{algpseudocode}
\usepackage[dvipsnames]{xcolor}
\usepackage{float}
\usepackage{amsthm}
\usepackage{caption}
\usepackage{subcaption}
\usepackage{multirow}
\usepackage{booktabs}
\usepackage{pifont}
\usepackage{makecell}

\newcommand{\eb}[1]{{\scriptsize\,$\pm$\,#1}}
\newtheorem{proposition}{Proposition}
\newcommand{\BPROP}{\begin{proposition}}   \newcommand{\EPROP}{\end{proposition}}

\newtheorem{corollary}{Corollary}[proposition]

\newtheorem{lem}{Lemma}
\newcommand{\BLEM}{\begin{lem}}   \newcommand{\ELEM}{\end{lem}}

\newcommand{\loss}{\mathcal{L}}
\newcommand{\D}{\mathcal{D}}

\newcommand{\til}[1]{\widetilde{#1}}
\newcommand{\grad}[1]{\nabla_{#1}}

\newcommand{\R}{\mathbb{R}}
\newcommand{\E}{\mathbb{E}}

\newcommand{\Mod}[1]{\ \mathrm{mod}\ #1}
\newcommand{\cmark}{\ding{51}}%
\newcommand{\xmark}{\ding{55}}%

\definecolor{commgreen}{RGB}{34,139,34}
\newcommand{\comm}[1]{\State \text{\textcolor{commgreen}{// #1}}}
\algrenewcommand\algorithmiccomment[1]{\hfill \textcolor{commgreen}{// #1}}

\DeclareMathOperator*{\argmin}{arg\,min} % Jan Hlavacek
\DeclareMathOperator*{\argmax}{arg\,max} % Jan Hlavacek

\algrenewcommand\algorithmicrequire{\textbf{Input:}}
\algrenewcommand\algorithmicensure{\textbf{Output:}}

\usepackage[pagebackref=true,breaklinks=true,letterpaper=true,colorlinks,bookmarks=false]{hyperref}
\usepackage{cleveref}
\crefname{subsection}{subsection}{subsections}

\definecolor{mydarkblue}{rgb}{0,0.08,0.45}
\definecolor{urlcolor}{rgb}{0,.145,.698}
\definecolor{linkcolor}{rgb}{.71,0.21,0.01}
\hypersetup{ %
	pdftitle={},  %TODO
	pdfauthor={},
	pdfsubject={}, %TODO
	pdfkeywords={},
	pdfborder=0 0 0,   
    pdfpagemode=UseOutlines,
	breaklinks=true, % so long urls are correctly broken across lines
	colorlinks=true,
	bookmarksnumbered=true,
	urlcolor=urlcolor,
	linkcolor=linkcolor,
	citecolor=mydarkblue,
	filecolor=mydarkblue,
	pdfview=FitH}
 \iccvfinalcopy % *** Uncomment this line for the final submission

\def\iccvPaperID{9426} % *** Enter the ICCV Paper ID here

\ificcvfinal\fi

\begin{document}

%%%%%%%%% TITLE
\title{Enhanced Meta Label Correction for Coping with Label Corruption}

\author{Mitchell Keren Taraday\\
Technion -- Israel Institute of Technology\\
{\tt\small mitchell@campus.technion.ac.il}
\and
Chaim Baskin\\
Technion -- Israel Institute of Technology\\
{\tt\small chaimbaskin@technion.ac.il}
}

\maketitle
\ificcvfinal\thispagestyle{empty}\fi

%%%%%%%%% ABSTRACT
\begin{abstract}
    Traditional methods for learning with the presence of noisy labels have successfully handled datasets with artificially injected noise but still fall short of adequately handling real-world noise. With the increasing use of meta-learning in the diverse fields of machine learning, researchers leveraged auxiliary small clean datasets to meta-correct the training labels.
    Nonetheless, existing meta-label correction approaches are not fully exploiting their potential. In this study, we propose an \textbf{E}nhanced \textbf{M}eta \textbf{L}abel \textbf{C}orrection approach abbreviated as EMLC for the learning with noisy labels (LNL) problem.
    We re-examine the meta-learning process and introduce faster and more accurate meta-gradient derivations. 
    We propose a novel teacher architecture tailored explicitly to the LNL problem,
    equipped with novel training objectives.
    EMLC outperforms prior approaches and achieves state-of-the-art results in all standard benchmarks.
    Notably, EMLC enhances the previous art on the noisy real-world dataset Clothing1M by $1.52\%$ while requiring $\times 0.5$ the time per epoch and with much faster convergence of the meta-objective when compared to the baseline approach.
    \footnote{Project page: \url{https://sites.google.com/view/emlc-paper}.}
\end{abstract}

%%%%%%%%% BODY TEXT

\section{Introduction}

\begin{figure}[t]
\includegraphics[width=0.5\textwidth]{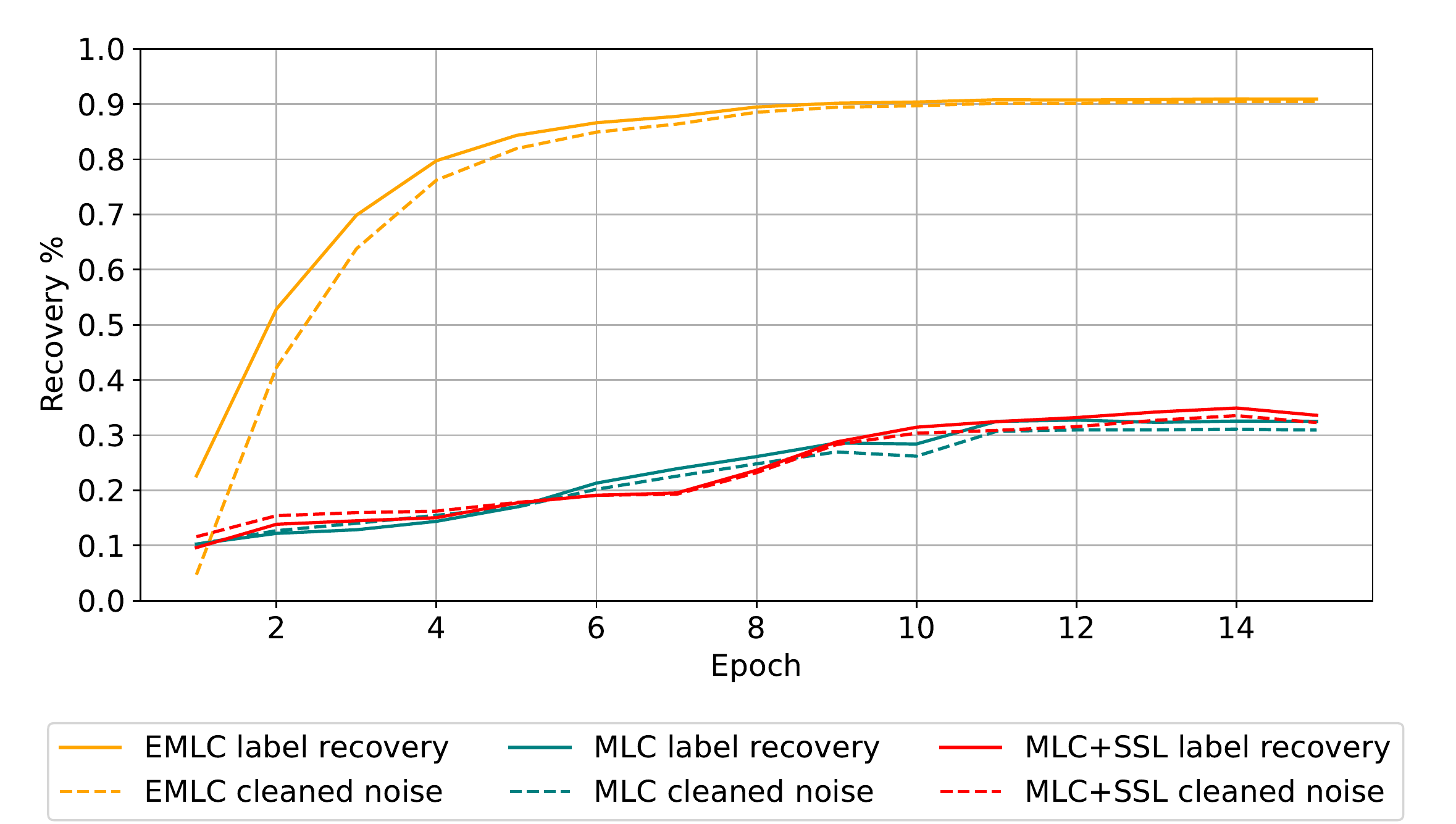}
\caption{Comparison of the training labels recovery (solid line) and the wrong labels recovery (dashed line) over training epochs on the CIFAR-10 dataset with $90\%$ symmetric noise.
EMLC (ours) is comapred to MLC \cite{mlc} and to MLC \cite{mlc} with self-supervised pretraining.
}
\label{fig:teaser}
\end{figure}

% Why LNL is important? applications, etc... [par 1]
The remarkable success of Deep Neural Networks (DNNs) for visual classification tasks 
is predominantly due to the availability of massive labeled datasets.
In many practical applications, obtaining the required amounts of reliable labeled data is intractable.
As a result, numerous works over the past decade have sought ways to reduce the amount of labeled data required for classification tasks.
Notably,
semi-supervised learning exploits unlabeled data \cite{uda,mpl},
transfer learning exploits prior knowledge obtained from different tasks \cite{classic-transfer,transfer-clip},
and self-supervised learning exploits data augmentations for label agnostic representation learning \cite{simclr,mae}.
Nonetheless, for many applications, the quality of the labeled data can be sacrificed for the sake of its quantity.
A prominent example is the process of crawling search engines and online websites as demonstrated by \cite{webvision, clothing}.
The crawling process is often easy to automate however results in significant amounts of label noise with complicated patterns.
% Consequently, learning from corrupted data has a huge potential for deploying large scale recognition algorithms.
A fundamental problem with such data, however, is that classical learning methodologies tend to fail when significant label noise is present \cite{overfitting}.
Therefore, designing learning frameworks that are able to cope with label corruption is a task of great importance.

While traditional methods for learning with noisy labels (LNL) are capable of handling data with immense artificial, injected noise, their ability to handle real-world noise remains highly limited. 
Thus, noisy labeled datasets alone limit the capability of learning from real-world noisy labeled datasets.
Luckily, in many real-world applications, while obtaining large
amounts of labeled data is infeasible, obtaining small amounts of labeled data is usually attainable.
Thus, the paramount objective is to find methods that can incorporate both large amounts of noisy labeled data and small amounts of clean data.
A natural choice is to adopt a meta-learning framework,
a popular design regularly used for solving various tasks.
Consequently, recent trends in LNL leverage meta-learning using auxiliary small clean datasets.
Prominent examples include meta-sample weighting \cite{mwn, metacleaner}, 
meta-robustification to artificially injected noise \cite{mnlt}, meta-label correction \cite{mlc}, and meta-soft label correction \cite{mslc}.

% While being remarkably successful, the two-models framework had one major disadvantage, which is the disregardment of the high-loss instances.
% As a result, multiple works tried to combine ideas from semi-supervised learning to exploit all the training data.
% Some of these works also improved the loss evaluation process, which is commonly referred to as \textit{noise separation} (Arazo \etal \cite{arazo}, Li \etal \cite{dividemix}).

In this work, we propose EMLC -- an enhanced meta-label correction approach for learning from label-corrupted data.
We first revise the bi-level optimization process by deriving a more accurate meta gradient used to optimize the teacher.
In particular, we derive an exact form of the meta-gradient for the one-step look-ahead approximation
and suggest an improved meta-gradient approximation for the multi-step look-ahead approximation.
We further provide an algorithm for efficiently computing our derivations using a modern hardware accelerator.
We empirically validate our derivations, demonstrating a significant improvement in convergence speed and training time.
We further propose a dedicated teacher architecture that
employs a feature extraction to generate initial predictions and incorporates them with the noisy label signal to generate refined soft labels for the student.
Our teacher architecture is completely independent of the student, avoiding the confirmation bias problem.
In addition, we propose a novel auxiliary adversarial training objective for enhancing the effectiveness of the teacher's label correction mechanism.
As demonstrated in \cref{fig:teaser}, our teacher proves to have a superior ability of purifying the training labels.

Our contributions can be summarized as follows:

\begin{itemize}
    \item We derive fast and more accurate procedures for computing the meta-gradient used to optimize the teacher.
    \item We design a unique teacher architecture in conjunction with a novel training objective toward an improved label correction process. 
    \item We combine these two components into a single yet effective framework termed EMLC. 
    We demonstrate the effectiveness of EMLC on both synthetic and real-world label-corrupted standard benchmarks.
\end{itemize}

% Meta Learning and its implications for LNL [par 3]
% Our suggestions [par 4]
% Our contributions [par 5]
\section{Related work}

The LNL problem was previously addressed by various approaches.
For instance, multiple works iteratively modified the labels to better align with the model's predictions \cite{reed, joint},
 estimated the noise transition matrix \cite{goldberger}, 
or applied a regularization \cite{phuber, mixup}.
The common pitfall of these methods is the phenomenon that DNNs tend to develop a confirmation bias, which causes the model to the confirm the corrupted labels.
To this end, Han \etal \cite{co-teaching} proposed a two-model framework
that employs a cross-model sample selection for averting the confirmation bias problem.
Being remarkably successful, many follow-up works tried to build on the two-model framework 
by either improving the sample selection process or 
 using semi-supervised tools to exploit the high-loss samples
and leveraging label-agnostic pretraining \cite{arazo,dividemix,c2d}.

Nevertheless, there is still a noticeable gap between the effectiveness of LNL methods and fully supervised training, especially on datasets with real-world noise.
In numerous practical scenarios, acquiring substantial quantities of labeled data may be impractical, but obtaining a small amount of labeled data is typically achievable. This observation motivates the use of meta-learning for tackling the LNL problem.
Consequently, multiple researchers have tried to exploit auxiliary small labeled datasets to apply meta-learning for LNL research.

Outside of LNL research,
meta-learning has begun to appear in diverse fields in machine learning,
including neural architecture search \cite{darts},
hyperparameter tuning \cite{hypertune},
few-shot learning \cite{maml}
and semi-supervised learning \cite{mpl}. 
Meta-learning often involves two types of objectives: an inner (lower-level) objective and a meta (upper-level) objective. 
Usually, meta-learning tasks are equipped with a large training set and a small validation set. 
The validation data (in many cases) is necessary for updating the meta-objective.

In the context of LNL, researchers leverage meta-learning to mitigate the effect of the label noise.
In particular, \cite{mwn, metacleaner} try to weight samples to degrade the effect of noisy samples on the loss.
MLNT \cite{mnlt} use a student--teacher paradigm and try to artificially inject random labels into
the data used to train the student model.
The student is encouraged to be close to the teacher model that does not observe the noisy data.
Meta-label correction \cite{mlc} and MSLC \cite{mslc} follow the student--teacher paradigm but, unlike MLNT \cite{mnlt},
use the teacher to produce soft training targets for the student.
The soft targets are produced in such a way that when the student is trained with respect to these targets, it performs well on the clean data.

EMLC builds upon the meta-label correction framework, which, as presented in prior work \cite{mlc,mslc}, does not appear to have fully exploited its potential.
We claim that the main pitfall of current label correction frameworks is that their teacher architecture is strongly entangled to the student, resulting in a severe confirmation bias.
On the contrary, EMLC is composed of a completely independent teacher, averting this problem.
In addition, EMLC leverages artificial noise injection for robust training.
Opposed to \cite{mnlt} however, we propose adversarial noise injection that we discover to be more effective for robustifying models against label noise.
Another major difference is that we use the artificial noise injection to better train the teacher whereas \cite{mnlt} use it to train the student.
\begin{figure}[t]
\includegraphics[width=0.4\textwidth]{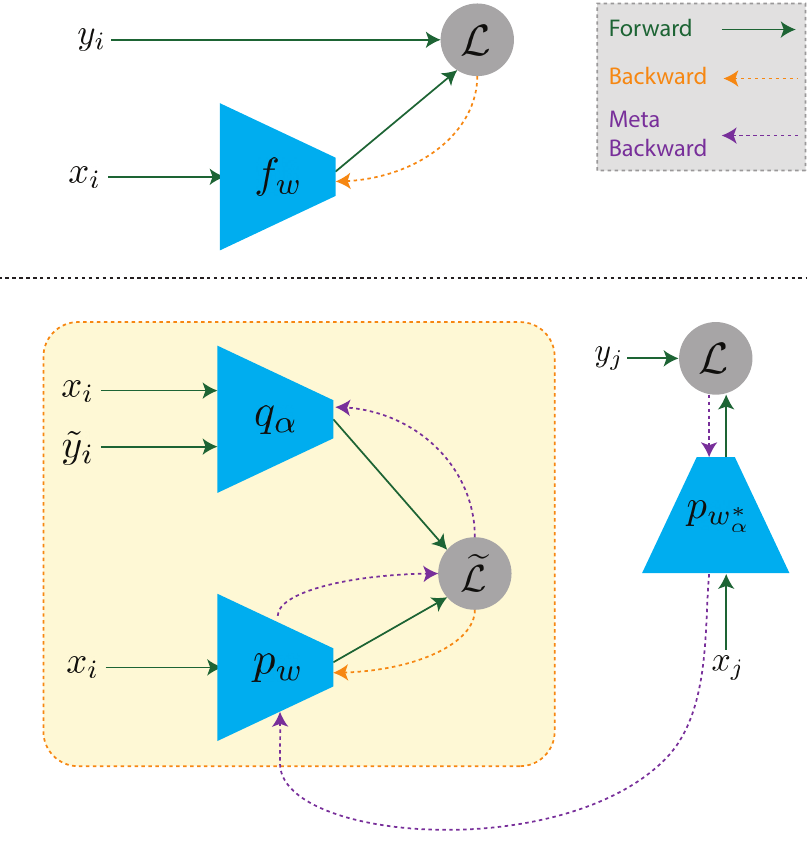}
\caption{Classical supervised learning via empirical risk minimization (upper part). 
Meta-learning for generating corrected labels via clean feedback optimization (lower part).
The meta-gradients are propagated through the optimization path of $p_w$ w.r.t. $w$.
The optimized student is denoted by $p_{w^*_\alpha}$}
\label{fig:erm_meta}
\end{figure}

\section{Methods} \label{sec:3}

In this section we discuss the distinct components of the proposed method.
We start by introducing the reader to the basic setting and assumptions in
LNL in \cref{formulation}.
In \cref{ermtometa}, we discuss the inability of classical learning algorithms
(i.e \textit{Empirical Risk Minimization} - ERM) to handle heavily corrupted data and look at how to exploit meta-generated labels as an alternative to the classical learning paradigm.
Accordingly, \cref{metabilevel} is devoted to bi-level optimization that repeatedly appears in meta-learning.
In \cref{metagrad}, we revisit meta-gradient approximation, which is the core of performing gradient-based bi-level optimization.
In addition, in \cref{sec:3.5}, we demonstrate how to compute the proposed derivations efficiently.
Finally, in \cref{teacher}, we elaborate on the proposed label correction architecture as well as on its training scheme.

\subsection{Learning with Noisy Labels Formulation} \label{formulation}

We initiate our discussion on noisy label classification by considering
a probability space $(\Omega, \mathcal{F}, p)$,
equipped with an instance random variable: $x: \Omega \rightarrow \mathcal{X}$,
along with a label and a (possibly) corrupted label random variable:  $y, \til{y}:  \Omega \rightarrow \mathcal{Y}$ where
$\mathcal{X}  \subseteq \R^d,\mathcal{Y} = \{1,...,C\}$ are the instance and label space, respectively.
We define the \textit{clean} distribution: $\D = (x,y)$
and the \textit{corrupted} distribution: $\til{\D} = (x,\til{y})$. 
The restriction on the corrupted distribution is that for each class $c \in \{1,...,C\}$,
the most probable corrupted label will match the true one: $c = \argmax_j p(\til{y} = j | y = c)$.
Given a family of parameterized predictors $\{ f_w: \mathcal{X} \rightarrow \mathcal{Y}\}$, the high level goal is to optimize the parameters $w$ to minimize the \textit{true expected risk}: $\mathcal{R}(f) = \E_{(x,y)}[\ell(f_w(x),y)]$ for some loss function $\ell: \mathcal{Y} \times \mathcal{Y} \rightarrow \R^+$ by accessing mostly $\til{\D}$.
In practice, $\til{\D}$ is most likely inaccessible and rather we are given i.i.d. samples of it $\{(x_i, \til{y}_i)\}_{i=1}^n$.
In the setting of interest, we are also supplied with a small set of \textit{clean} i.i.d. samples
$\{(x_j, y_j)\}_{j=1}^m$ where $m \ll n$.

\subsection{From ERM to Meta-Learning} \label{ermtometa}

According to the ERM principle in supervised learning,
$f_w$ should be optimized by considering the \textit{empirical risk}:

\begin{align}
    w^* = \argmin_w \frac{1}{n} \sum_{i=1}^n \ell(f_w(x_i),y_i)
\end{align}

To make optimization feasible, $\ell$ is chosen to be a differentiable function.
In particular, if we generalize $f_w$ to model a probability distribution, $p_w(y|x)$, 
 the \textit{cross-entropy} loss \cite{celoss} is most commonly used:

\begin{align}
    w^* = \argmin_w \frac{1}{n} \sum_{i=1}^n CE(y_i,p_w(y|x_i))
\end{align}

In our case, the labels $\til{y}_i$ are possibly corrupted. 
Therefore, applying the ERM principle would give rise to noisy label memorization (due to the universal approximation theorem)
leading to a severe degradation in the obtained predictor's performance \cite{elr,overfitting}.
This problem is addressed by replacing $y_i$ with a label correction architecture $q_\alpha(y|x_i, \til{y}_i)$, which will produce corrected (soft) labels for training $p_w$.
For the ease of reading, from this point on, we refer to $p_w$ as the \textit{student} architecture and $q_\alpha$ as the \textit{teacher} architecture.
Recall that the real goal is to optimize the student to minimize the true expected error.
Hence, given a small set of clean samples, the teacher should produce soft labels in a way that would cause the optimized student to have small empirical risk on the clean set, as demonstrated in \cref{fig:erm_meta}. 
Therefore, the objective may be formalized as the following bi-level optimization problem:
\begin{align}
    & \min_\alpha \loss(w^*(\alpha))\\
    \label{eqn:lowerlevel}
    & s.t \text{  } w^* = \argmin_w \til{\loss}(w,\alpha)
\end{align}
where:
\begin{align}
    \label{eqn:cleanlossdef}
    & \loss(w^*(\alpha)) = \frac{1}{m} \sum_{j=1}^m CE(y_j,p_{w^*(\alpha)}(y|x_j)) \\
    \label{eqn:noisylossdef}
    & \til{\loss}(w, \alpha) = \frac{1}{n} \sum_{i=1}^n CE(q_{\alpha}(y|x_i, \til{y}_i),p_w(y|x_i))
\end{align}

Note that under relatively mild conditions, $w^*$ can be written (locally) as a function of $\alpha$. 
This can be observed from stationary conditions applied to \cref{eqn:lowerlevel}: $\grad{w} \til{\loss}(w,\alpha) = 0$ and applying the implicit function theorem.

\subsection{Bi-Level Optimization} \label{metabilevel} 
Bi-level optimization problems appear in diverse sub-fields in machine learning and in meta-learning in particular \cite{darts,hypertune,maml,mpl}.
Such problems are composed of an outer level optimization problem that is usually constrained by another inner level problem.
We relate to the upper level problem parameters as the \textit{meta-parameters} and to the parameters in the lower level as the \textit{main parameters}. 
The fact that each meta-parameter value defines a new lower level problem makes applying gradient-based optimization methods to such situations particularly difficult due to the iteration complexity. 
A solution to this problem is to approximate $w^*$ with $k$-step look-ahead SGD,
optimizing both the \textit{main} and  \textit{meta}- parameters simultaneously.
The joint optimization process can be summarized in \cref{alg:kstep}.

\begin{algorithm}
\caption{Bi-Level Optimization via $k$-step SGD Look-ahead Approximation}\label{alg:kstep}
\begin{algorithmic}[1]
\Require Clean and noisy datasets $\D, \til{\D}$,
number of training steps $T$,
initial parameters $w^{(0)},\alpha^{(0)}$,
learning rates $\eta_w, \eta_\alpha$.
\Ensure Optimized parameters $w^{(T)},\alpha^{(T)}$.

\For{$t = 0$ ,..., $T-1$}
        \comm{Sample clean and noisy batches from the datasets}
        \State {$\mathcal{B} , \til{\mathcal{B}} \gets Sample(\D), Sample(\til{\D})$}
        \comm{Update $w$ by descending $\til{\loss}(w,\alpha)$ w.r.t $w$}
        \State {$w^{(t+1)} \gets w^{(t)} - \eta_w \grad{w} \til{\loss}(w, \alpha^{(t)})\bigg\rvert_{w=w^{(t)}}$}
        \If{$t \Mod k = k-1$}
            \comm{Unroll $w^{(t+1)}(\alpha)$ to compute the meta gradient} 
            \State {$g_\alpha = \text{\texttt{META\_GRAD}}(\loss(w^{(t+1)}), \alpha)$}
            \comm{Update $\alpha$ by descending $\loss(w^{(t+1)}(\alpha))$ w.r.t $\alpha$}
            \State {$\alpha^{(t+1)} \gets  \alpha^{(t)} - \eta_\alpha g_\alpha$}
        \Else
            \State{$\alpha^{(t+1)} \gets \alpha^{(t)}$}
        \EndIf
    \EndFor

\end{algorithmic}
\end{algorithm}

During training, the higher order dependencies of $w$ on $\alpha$ are neglected,
thereby allowing the meta-gradient to be computed, as we discuss in the next section.

\subsection{Revisiting the Meta-Gradient Approximation} \label{metagrad}

As opposed to Meta-Label Correction \cite{mlc}, we propose better approximations for both the single and multi-step meta-gradients when updating the teacher.
In the next section, we establish an efficient method for carrying out the meta-gradient computation.

We explicitly differentiate between the cases $k=1$ and $k>1$.
We begin by deriving the \textit{one-step approximation meta-gradient}.

\BPROP
\label{prop:os}
(One-Step Meta-Gradient) 
Let $g_w:= \grad{w} \loss(w^{(t+1)})$
where $w^{(t+1)}$ are the student's new parameters obtained when last updated. 
Let $H_{w \alpha}^{(t)} := \grad{w \alpha} \til{\loss}(w^{(t)}, \alpha^{(t)})$ where $w^{(t)}$ are the student's original parameters (before its update).
Let $\alpha^{(t)}$ be the teacher's parameters before its update. 
Then, the one-step meta-gradient at time $t$ denoted by $\grad{\alpha}^{(t)}$ is given by:
\begin{align}
    [\grad{\alpha}^{(t)}]^T = -\eta_w g_w^T  H_{w \alpha}^{(t)}
\end{align}
\EPROP

The proof of \cref{prop:os} is given in \cref{proof:single-step}.
We proceed by considering the case of $k > 1$.
In this case, we do not compute the exact meta-gradient. Instead, we approximate it.

\BPROP
\label{prop:ms}
($k$-Step Meta-Gradient Approximation with Exponential Moving Average of Mixed Hessians) 
Assume that $t \Mod k = k-1$ and let $g_w:= \grad{w} \loss(w^{(t+1)})$
where $w^{(t+1)}$ is the student's new parameters obtained when last updated. 
Consider the mixed Hessians in the last $k$ steps: $H_{w \alpha}^{(\tau)} :=  \grad{w \alpha} \til{\loss}(w^{(\tau)}, \alpha^{(\tau)})$
where $t-k+1 \leq \tau \leq t$.
Let $\alpha^{(t)} = \alpha^{(t-1)} = ... = \alpha^{(t-k+1)}$ be the teacher's parameters before its update. 
Then, the $k$-step meta-gradient denoted by $\grad{\alpha}$ can be approximated by:
\begin{align}
    [\grad{\alpha}^{(t)}]^T \approx -\eta_w g_w^T  \sum_{\tau=t-k+1}^{t} \gamma_w^{t-\tau} H_{w \alpha}^{(\tau)} 
\end{align}
where $\gamma_w = 1 - \eta_w$.
\EPROP
The proof of \cref{prop:ms} is given in \cref{proof:multi-step}.

\subsection{Computational Efficiency of the Meta-Gradient} \label{sec:3.5}

\begin{figure*}
\centering
\includegraphics[width=0.75\textwidth]{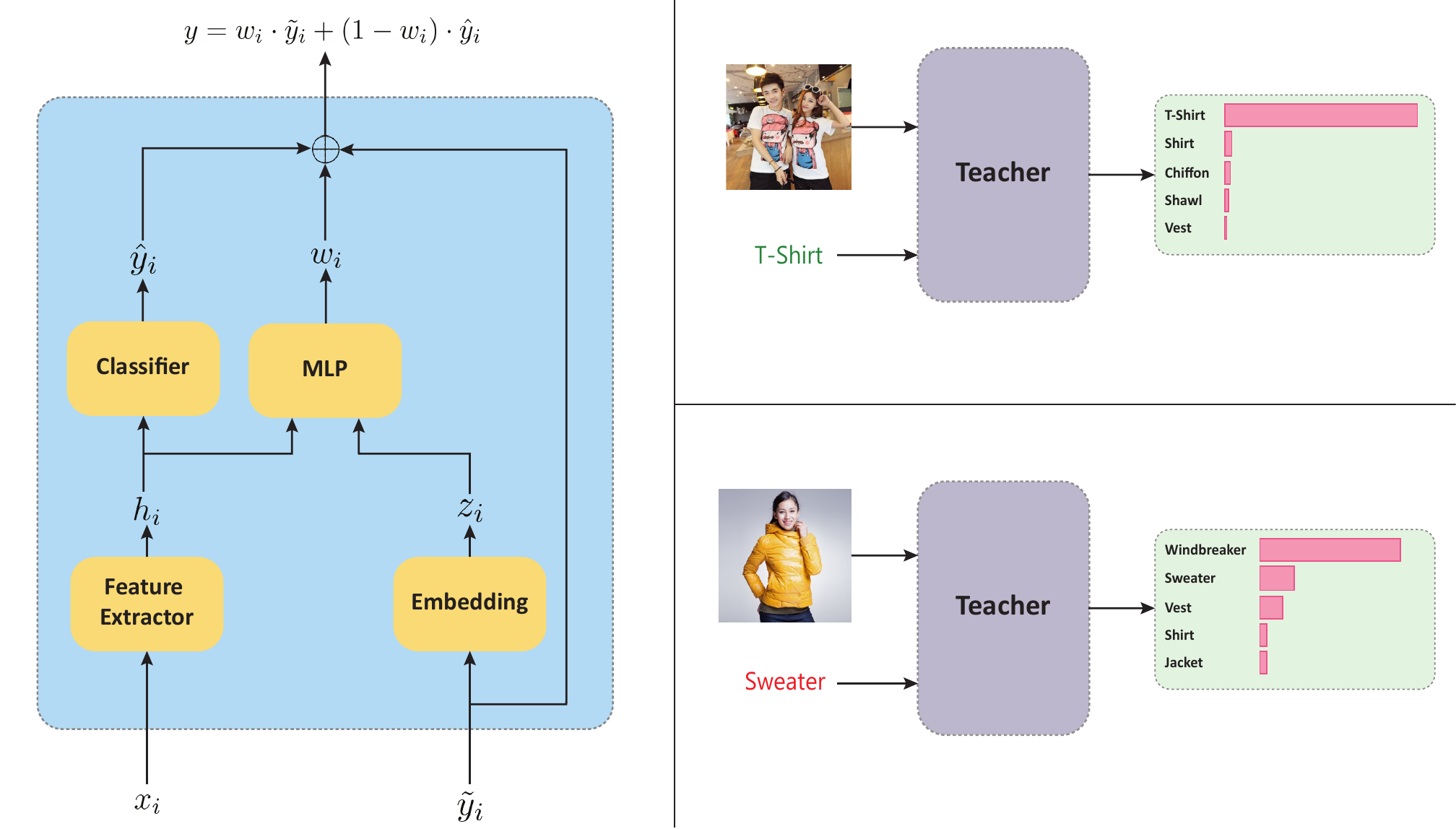}
\caption{The proposed teacher architecture (left) and typical teacher's output distributions when fed with clean/noisy sample (right).  
In our architecture, the sample is initially fed to a feature extractor to obtain a representation $h_i$ and then to a classifier to obtain an initial prediction $\hat{y}_i$.
The (possibly) noisy label is then fed to an embedding layer to obtain a label embedding $z_i$.
Both of the embeddings are then concatenated and are fed to an MLP which produces a weight $w_i \in (0,1)$. Finally, $w_i$ is used to gate the initial prediction and the noisy label to produce the final prediction.
}
\label{fig:st}
\end{figure*}

Calculating the one-step meta-gradient as in \cref{prop:os} or
the multi-step meta-gradient as in \cref{prop:ms}
in a straightforward manner is inefficient,
as it requires computing higher-order derivatives of large DNNs. 
In this section, we provide a few workarounds that allow us to carry out the computations efficiently.
Our first observation is that the sample-wise mixed Hessian can be expressed as a product of two first-order Jacobians. Formally:

\BPROP
\label{prop:jacobians}
Let us consider the sample-wise lower loss: 
$\til{\ell}_i(w,\alpha) := CE(q_{\alpha}(y|x_i, \til{y}_i),p_w(y|x_i))$. 
Recall from \cref{eqn:noisylossdef} that 
$\til{\loss}(w, \alpha)  = \frac{1}{n} \sum_{i=1}^n \til{\ell}_i(w,\alpha)$. 
Denote by $J_w(i):= J_w(\log(p_w(y|x_i)))$ and by
$J_\alpha(i) := J_\alpha(q_{\alpha}(y|x_i, \til{y}_i))$.
We claim that:
\begin{align}
    \grad{w \alpha} \til{\ell}_i(w,\alpha) = 
    -[J_w(i)]^T[J_\alpha(i)]
\end{align}
\EPROP

The proof of \cref{prop:jacobians} is given in \cref{proof:jacobians}.

\begin{corollary}
    \label{cor:metagrad}
    Using \cref{prop:os}, the one-step meta-gradient can be expressed as:
    \begin{align}
        [\grad{\alpha}^{(t)}]^T =
        \frac{\eta_w}{n} \sum_{i=1}^n g_w^T  [J_{w^{(t)}}(i)]^T[J_{\alpha^{(t)}}(i)]
    \end{align}
    Using \cref{prop:ms} the k-step meta-gradient approximation can be expressed as:
    \begin{align}
        [\grad{\alpha}^{(t)}]^T =
        \frac{\eta_w}{n} \sum_{i=1}^n \sum_{\tau=t-k+1}^{t}
        \gamma_w^{t-\tau} g_w^T  [J_{w^{(\tau)}}(i)]^T[J_{\alpha^{(t)}}(i)]
    \end{align}
\end{corollary}

\begin{table*}[t]
    	\centering
	\begin{tabular}	{l | ccccc | cccc } 
        \toprule
        \multicolumn{1}{l|}{\textbf{Dataset}} &
        \multicolumn{5}{c|}{\textbf{CIFAR-10}} &
        \multicolumn{4}{c}{\textbf{CIFAR-100}} \\
        \midrule
        \textbf{Method} &
        \textbf{20\%} & \textbf{50\%} & \textbf{80\%} & \textbf{90\%} & \textbf{Asym. 40\%} &
        \textbf{20\%} & \textbf{50\%} & \textbf{80\%} & \textbf{90\%} \\
        \midrule \midrule
        Cross-Entropy & 86.8 & 79.4 & 62.9 & 42.7 & 83.2
        & 62.0 & 46.7 & 19.9 & 10.1\\
        % Co-Teaching+ \cite{co-teaching} & 89.5 & 85.7 & 67.4 & 47.9 & -- & 
        % 65.6 & 51.8 & 27.9 & 13.7 & -- \\
        % MixUp \cite{mixup} & 95.6 & 87.1 & 71.6 & 52.2 & --
        % & 67.8 & 57.3 & 30.8 & 14.6 & -- \\
        % M-Correction \cite{arazo} & 93.8 & 91.9 & 86.6 & 68.7 & 87.4
        % & 73.9 & 66.1 & 48.2 & 24.3 & -- \\
        % DivideMix \cite{dividemix} & 95.7 & 94.4 & 92.9 & 75.4 & 92.1
        % & 77.3  & 74.6 & 60.2 & 31.5 & 72.1 \\
        % ELR+ \cite{elr} & 95.8  & 94.8  & 93.3  & 78.7 & 93.0
        % & 77.6  & 73.6 & 60.8 & 33.4 & 77.5 \\
        % AugDesc \cite{augmentations} & 96.3 & 95.4 & 93.6 & 91.8 & 94.3
        % & 79.5 & 77.5 & 66.0 & 40.9 & -- \\
        % C2D \cite{c2d} & 96.2 &  95.1 & 94.3 & 93.4 & 90.8
        % & 80.89 & 79.20 & 71.53 & 63.91 & 77.78\\
        % \midrule
        MW-Net \cite{mwn} & 89.76 & -- & 56.56 & -- & 88.69 & 
        66.73 & -- & 19.04 & -- \\
        MLNT \cite{mnlt} & 92.9  & 88.8  & 76.1  & 58.3  & 88.6
        & 67.7 & 58.0 & 40.1 & 14.3 \\
        MLC \cite{mlc} & 92.6 & 88.1 & 77.4 & 67.9 & --
        & 66.5 & 52.4 & 18.9 & 14.2 \\
        MSLC \cite{mslc} & 93.4 & 89.9 & 69.8 & 56.1 & 91.6 & 
        72.5 & 65.4 & 24.3 & 16.7 \\
        FasTEN \cite{fasten} & 91.94 & 90.07 & 86.78 & 79.52 & 90.43
        & 68.75 & 63.82 & 55.22 & -- \\
        \midrule
        EMLC ($k=1$)
        % CIFAR-10
        & \begin{tabular}{c}91.8  \\\eb{0.51}\end{tabular}  %92.03, 91.22, --- 92.15, 
        & \begin{tabular}{c}91.16  \\\eb{0.25}\end{tabular} %90.91, 91.15, --- 91.41, 
        & \begin{tabular}{c}\textbf{90.95}  \\ \textbf{{\eb{0.24}}} \end{tabular} %90.62, 91.13 --- 91.11,
        & \begin{tabular}{c}\textbf{90.71} \\ \textbf{\eb{0.07}} \end{tabular} %90.74, 90.63 --- 90.77
        & \begin{tabular}{c}\textbf{91.81}  \\ \textbf{{\eb{0.05}}} \end{tabular} %91.75, 91.83 --- 91.85
        % CIFAR-100
        & \begin{tabular}{c}72.48  \\ \eb{0.44} \end{tabular} % 72.39, 72.10, 72.96
        &  \begin{tabular}{c}67.08 \\ \eb{0.35} \end{tabular} % 65.91, 68.32, 67.03
        & \begin{tabular}{c}\textbf{60.37}  \\ \textbf{{\eb{0.31}}} \end{tabular} %60.71, 60.29, 60.11
        & \begin{tabular}{c}\textbf{54.04}  \\ \textbf{{\eb{0.57}}} \end{tabular} % 53.64 , 54.69,  53.79
        \\
        EMLC ($k=5$)
        & \begin{tabular}{c}\textbf{93.53}  \\ \textbf{{\eb{0.21}}} \end{tabular} %93.74, 93.53, --- 93.33
        & \begin{tabular}{c}\textbf{92.63}  \\ \textbf{{\eb{0.05}}} \end{tabular} %92.69, 92.56, --- 92.63
        & \begin{tabular}{c}89.89  \\ \eb{0.07} \end{tabular} %89.84, 89.99, --- 89.85
        & \begin{tabular}{c}89.57  \\ {\eb{0.30}} \end{tabular} %90.35, 90.16, --- 88.21
        & \begin{tabular}{c}\textbf{91.82}  \\ \textbf{\eb{0.19}} \end{tabular} %92.00, 91.89, --- 91.56
        % CIFAR-100
        & \begin{tabular}{c}\textbf{73.05}  \\ \textbf{{\eb{0.20}}} \end{tabular} % 73.27, 72.97, 72.90
        & \begin{tabular}{c}\textbf{68.61}  \\ \textbf{{\eb{0.34}}} \end{tabular} % 68.98, 68.32, 68.53
        & \begin{tabular}{c}\textbf{60.51}  \\ \textbf{{\eb{0.19}}} \end{tabular} % 60.26, 60.71, 60.58
        & \begin{tabular}{c}52.49  \\ \eb{0.18} \end{tabular} % 52.40, 52.74, 52.33
        \\
		\bottomrule
	\end{tabular}
	\caption
		{
		%\small	
		Comparison with state-of-the-art methods in test accuracy (\%) on CIFAR-10 and CIFAR-100 datsets
        corrupted with multiple levels and types of noise.
        The reported standard deviations are based on $5$ runs using different seeds for each setting.
        }
	\label{tbl:cifar}
\end{table*}

\cref{cor:metagrad} can be exploited to compute the meta-gradient efficiently as demonstrated in \cref{alg:fpmg}.

\begin{algorithm}[t]
\caption{Fast and Precise Meta Gradients (FPMG)}\label{alg:fpmg}
\begin{algorithmic}[1]
\Require Student weights obtained from the inner problem iterations $w^{(t-k+1)},...,w^{(t+1)}$,
last teacher weights $\alpha^{(t)}$.
\Ensure Meta-gradient for the current iteration $\grad{\alpha}^{(t)}$.

\State {$g_w \gets \grad{w} \loss(w^{(t+1)})$} \Comment{Student feedback gradient}
\State{$\grad{\alpha}^{(t)}, d_w \gets 0_\alpha, 1$}
\For{$\tau = t,...,t-k+1$}
    \For{$i \in \til{\mathcal{B}}$}
        \comm{Compute JVP using forward mode AD}
        \State {$jvp_i \gets [J_{w^{(t)}}(i)] g_w$}
        \comm{Compute VJP using backward mode AD}
        \State {$res_i \gets (jvp_i)^T [J_{\alpha^{(t)}}(i)]$}
    \EndFor
    \comm{Accumulate the contribution of $\tau$ to $\grad{\alpha}^{(t)}$}
    \State{$\grad{\alpha}^{(t)} \gets \grad{\alpha}^{(t)} + d_w \frac{\eta_w}{n} \sum_{i=1}^n res_i$}
    \State{$d_w \gets (1-\eta_w)d_w$} \Comment{Update discount factor}
\EndFor

\end{algorithmic}
\end{algorithm}

% In the case of a one-step meta-gradient, one can 
% compute $g_w$ once via a single backward mode AD (automatic differentiation). 
% The intermediate result can be stored on an intermediate variable $g$.
% Then, for each $i$ (in a parallelized fashion), one can use $g$ to compute $g_w^T [J_{w^{(t)}}(i)]^T$ which is a transposed
% Jacobian-vector product (JVP) that can be computed efficiently via a forward mode AD.
% This intermediate result can be stored on an intermediate variable $jvp_i$.
% Finally, $jvp_i$ can be used to compute $jvp_i^T [J_{\alpha^{(t)}}(i)]$, which can be computed using a single backward mode AD.
% All in all, the single-step meta-gradient can be computed using two backward mode ADs and one forward mode AD.

In terms of computation, the FPMG algorithm can be easily batchified and hence takes roughly only three times the computation required for a single backward mode AD used in a simple lower level update. 
In terms of memory (in addition to the original and updated student parameters), only the intermediate variables that accumulate for a one-student gradient and an $n \times C$ matrix for storing the JVPs for all the samples, must be stored, which is very efficient.

% For computing the $k$-step meta-gradient, step $1$ in the above process is performed once and steps $2$ and $3$ are repeated $k$ times, yielding  a $2k+1$ multiplicative factor for the computation.

Since the computation grows linearly with the number of batches, the only bottleneck in the approach is the memory required to store all the $k+1$ versions of the student's parameters obtained in its optimization process. This is a standard snag in such bi-level optimization problems \cite{maml, bilevel}.
A summary of the differences of FPMG from MLC's \cite{mlc} meta gradient computation is given in \cref{appendix:meta_grad_comp}.

\subsection{Teacher Architecture and Optimization} \label{teacher}

Recall that the key to applying meta-learning to the LNL problem lies in the \textit{teacher} architecture.
The teacher $q_\alpha(y|x_i, \til{y}_i)$ aims to correct the labels of (possibly) corrupted sample--label pairs,
given the sample and its corresponding (possibly) corrupted label. 
Thus, it is important to design an appropriate teacher architecture.
We propose a teacher architecture which is especially tailored for correcting corrupted labels.
In contrast to MLC \cite{mlc} and MSLC \cite{mslc}, our teacher architecture possesses its own feature extractor and a classifier to produce its own predictions.
As a result, the proposed label correction procedure is completely disentangled from the student. This structure decreases noisy label memorization and confirmation bias. 
In addition, the proposed architecture carefully takes advantage of the noisy label signal.
The noisy label is used to extract label embedding, which is then combined with the image embedding and  fed to an MLP with a single output neuron.
The MLP learns to gate the teacher's predictions and the input noisy label so that 
when the input label is not actually corrupted, the input label signal would be preferred.
Our proposed teacher architecture is summarized in \cref{fig:st}.
In addition to the meta-learning optimization objective,
the unique structure of the proposed architecture enables us to further leverage the extra clean data to design additional supervised learning objectives for training the teacher.
We first notice that both the feature extractor and the classifier can be trained by considering simple supervised \textit{cross entropy} loss on the clean training data.
Followed by the findings in \cite{augmentations}, we apply AutoAugment data augmentation \cite{autoaugment} on the extra clean data.
Nonetheless, the label embedding layer and the label-retaining MLP predictor are omitted from this loss.
To this end, we propose to artificially inject noise into a portion of the clean training data and encourage the MLP to predict whether the input label is clean or corrupted.
In practice, we corrupt half of each clean training batch and use binary cross-entropy loss at the end of the MLP.
We propose two corruption strategies; 
in the \textit{random corruption} strategy, we sample uniformly i.i.d. random labels and assign them to half of the batch; 
in the \textit{adversarial corruption} strategy, we corrupt an image's label
with the strongest incorrect prediction obtained by the teacher's classifier.
We hypothesize that the adversarial corruption strategy would cause the MLP to identify clean labels when the teacher's predictions are uncertain.
The final teacher training objective can be summarized as follows:

\begin{align}
    \loss(\alpha) = \loss_{CE}(\alpha) + \loss_{BCE}(\alpha) + \loss_{META}(w^*(\alpha)) 
\end{align}

\section{Experiments}

In this section, we extensively verify the empirical effectiveness of EMLC, both qualitatively and quantitatively.

In \cref{sec:cifar,sec:clothing} we 
verify the empirical effectiveness of EMLC on the three major benchmark datasets used in the literature \cite{mwn, mlc, mnlt, fasten, mslc} about meta-learning for LNL.
We inject synthetic random noise at multiple levels and of assorted types into the CIFAR-10 and CIFAR-100 datasets \cite{cifar}, which are correctly labeled datasets. 
On the other hand, the Clothing1M dataset \cite{clothing} is a massive dataset collected from the internet, containing many mislabeled examples.
In all our experiments on the benchmark datasets we validate EMLC with both
the single-step look-ahead optimization strategy and the multi-step optimization strategy. 

In \cref{sec:meta} we discuss the efficiency of the proposed meta-learning procedures used in EMLC in terms of computation time and speed of convergence.

\begin{table}[t]
    	\centering
    \setlength\extrarowheight{2pt}
	\begin{tabular}	{l c c }
        \toprule	 	
			\textbf{Method} & \textbf{Extra Data} & \textbf{Test accuracy} \\
        \midrule
			Cross-entropy & \xmark & 69.21 \\		
			Joint-Optim \cite{jointoptim}  & \xmark & 72.16\\			
			P-correction \cite{pcorrection} & \xmark & 73.49\\
            C2D \cite{c2d} & \xmark & 74.30\\
			DivideMix \cite{dividemix} & \xmark & 74.76\\
			ELR+ \cite{elr} & \xmark & 74.81\\
            AugDesc \cite{augmentations} & \xmark & 75.11\\
            SANM \cite{sanm} & \xmark & 75.63 \\
        \midrule
            Meta Cleaner \cite{metacleaner} & \cmark & 72.50 \\ 
            Meta-Learning \cite{mnlt}  & \cmark & 73.47 \\	
            MW-Net \cite{mwn} & \cmark & 73.72 \\
            FaMUS \cite{famus} & \cmark & 74.40 \\
            MLC \cite{mlc} & \cmark & 75.78\\
            MSLG \cite{MSLG} & \cmark & 76.02 \\
            Self Learning \cite{self-learning} & \cmark & 76.44 \\
            FasTEN \cite{fasten} & \cmark & 77.83 \\
            \midrule
            EMLC ($k=1$) & \cmark & \textbf{79.35} \\
            EMLC ($k=5$) & \cmark & 78.16 \\
		\bottomrule
	\end{tabular}
	\caption
		{
		%\small	
		Comparison with state-of-the-art methods in test accuracy (\%) on Clothing1M. 
        The methods in the lower part of the table use extra clean data,
        while the methods in the upper part do not (above an below the dividing line).
        }
	\label{tbl:clothing}
\end{table}

 In \cref{sec:ablations} we perform ablation studies on the distinct components of EMLC and on the number of look-ahead steps.
 % In the first study we examine how EMLC performs by gradually removing some of its components, justifying the essence of each component.
 % In addition, we examine how EMLC performs with different numbers of look-ahead steps, demonstrating the robustness of EMLC to varying numbers of look-ahead steps.

\subsection{CIFAR-10/100} \label{sec:cifar}

CIFAR-10 is a $10$ class dataset consisting of $50$k training and $10$k testing RGB tiny images.
Likewise, CIFAR-100 is a $100$ class dataset with images of the same dimensionality and the same total amount of training and testing images. 

To evaluate the effectiveness of the EMLC framework, we follow previous works on meta-learning for LNL \cite{mwn,mnlt,fasten,mslc} and adopt the standard protocol for validating our framework.
For the small clean dataset, $1,000$ samples are randomly selected and separated from the training data. 
The small clean dataset is balanced, yielding  
$100$ instances from each category in the CIFAR-10 dataset
and $10$ samples from each category in the CIFAR-100 dataset.
The rest of the training data is designated as the "large" noisy dataset
and is injected with artificial label noise.

Two distinct settings for the noise injection are used: \textit{Symmetric} and \textit{Asymmetric} noise.
As these are standard in LNL research, we refer the reader to \cref{appendix:sym_vs_asym} for more details on these settings.

Following prior work, in the CIFAR-10 experiments we use ResNet-34 architecture \cite{resnet} whereas in the CIFAR-100 experiments we use ResNet50 architecture.
In both experiments we initialize the model using SimCLR \cite{simclr} self-supervised pretraining, following \cite{c2d}.
We note that owing to self-supervised pretraining, only $15$ epochs are needed for obtaining optimal results. We report our results on the CIFAR datasets with $\{20\%, 50\%, 80\%, 90\%\}$ symmetric noise and $40\%$ asymmetric noise in CIFAR-10.
We treat the clean data as a validation set,
and report the results using the student model when its validation score is the highest.
We compare our results with previous methods in \cref{tbl:cifar}.

EMLC proves to be very effective, obtaining state-of-the-art results in all the experiments. 
Notably, EMLC superbly surpasses prior works when facing high noise levels, maintaining an accuracy of more than $90\%$ in CIFAR-10 and more than $50\%$ in CIFAR-100 at a $90\%$ noise rate.
Furthermore, EMLC shows an impressive improvement over the state-of-the-art methods of $2.16\%$ on CIFAR-10 and $3.58\%$ on CIFAR-100 in  medium-level $50\%$ noise setting.
We observe that although both the one-step and multi-step strategies are superior to prior methods, the multi-step strategy worked the best.
Regarding the number of steps chosen for the multi-step strategy, in our experiments we found that $k=5$ was optimal.
In \cref{sec:ablations} we show that $k=5$ is indeed optimal for the $50\%$ noise rate in CIFAR-100.

\subsection{Clothing1M} \label{sec:clothing}

\begin{figure*}[ht]
\centering
\includegraphics[width=0.85\textwidth]{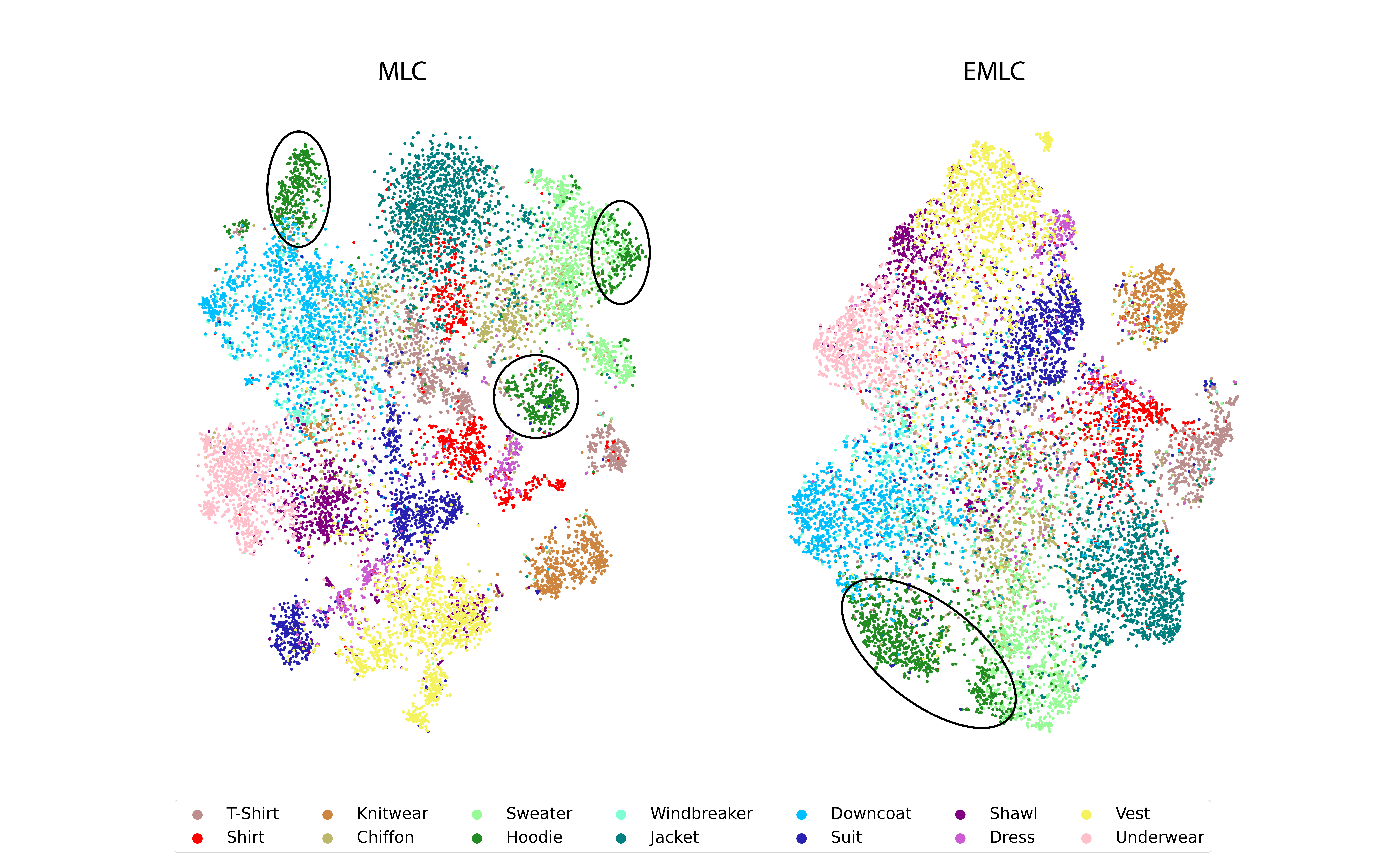}
\caption{Comparison of the tSNE plot of the validation set embeddings obtained by the student when trained on Clothing1M.
}
\label{fig:visualization}
\end{figure*}

Clothing1M is a large-scale clothing dataset obtained by crawling images from several online shopping websites. 
The dataset consists of $14$ classes.
The dataset contains more than a million noisy labeled samples as well as a small set (around $50K$) of clean samples and small validation and testing datasets.
We follow prior works on LNL \cite{dividemix, augmentations, mnlt} and use ResNet-50 architecture \cite{resnet} pretrained on ImageNet.
We train both architectures for $5$ epochs, using both the single-step and multi-step look-ahead strategies.
Additional experimental details can be found in \cref{appendix:implementation}.
As in the CIFAR experiments, we report the results using the student model with the highest validation score.
We compare our results with previous approaches in \cref{tbl:clothing}.
As can be observed, our method reaches a clear-cut result on the challenging Clothing1M dataset that surpasses the state-of-the-art methods by $1.52\%$.
In addition, the results show a significant improvement of over $3.5\%$ to MLC \cite{mlc}.
In contrast to our CIFAR experiments, in the Clothing1M experiment, the one-step meta-gradient proved to be better than the multi-step meta-gradient, showing the essence of  both strategies. 
Qualitatively, we visually compare the validation set representations of the trained student model of EMLC and MLC\cite{mlc} using a t-SNE \cite{tsne} plot in \cref{fig:visualization}.
As can be observed, EMLC manages to keep the same categories clustered whereas MLC \cite{mlc} fails to do so on multiple categories.

\subsection{Meta-Learning} \label{sec:meta}

We now discuss the empirical effectiveness of our meta-gradient approximation and computation, demonstrated on the Clothing1M dataset.

\begin{figure}
\centering
\includegraphics[width=0.50\textwidth]{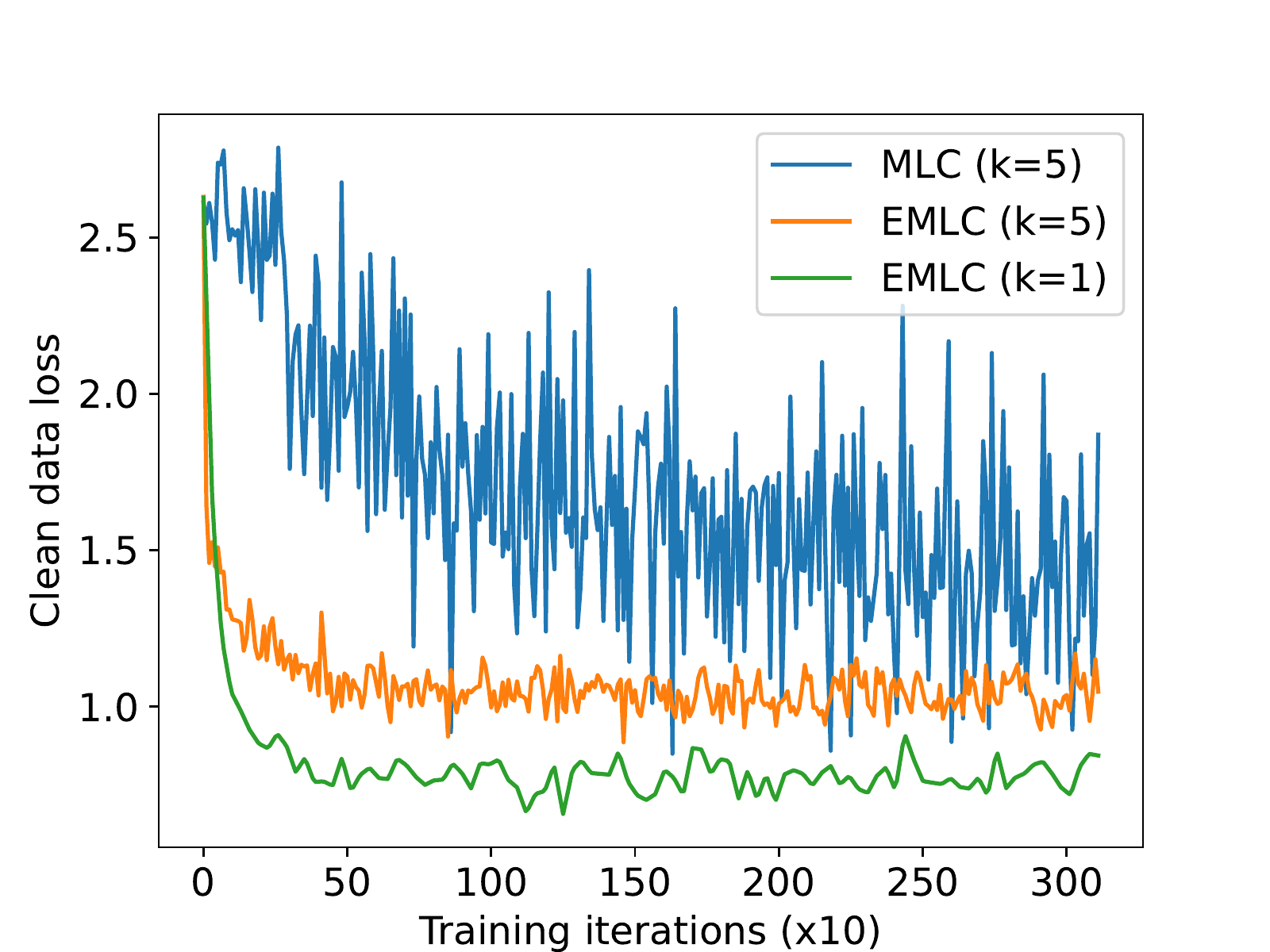}
\caption{The meta-objective value for the first two epochs.
The horizontal axis indicates the training iterations.
}
\label{fig:losses}
\end{figure}

In terms of speed of convergence, we compare the convergence, in  MLC \cite{mlc} and  EMLC, of the student's meta-evaluation loss . 
Recall that the meta-learning objective is to reduce this loss.
The loss dynamics are presented in \cref{fig:losses}.
Note that the meta-regularization loss of EMLC converges in the first two epochs for both $k=1$ and $k=5$. 
In contrast, MLC's \cite{mlc} regularization loss has yet to converge at this point and has a much noisier optimization process.
In addition, it can be observed that our method (for both $k=1$ and $k=5$)
has a lower clean feedback loss compared to MLC.
In the Clothing1M experiment, however, the single-step optimization process dominates the multi-step optimization process in terms of the final loss value, which correlates with the final accuracy.

In terms of computation, we compare the time required by EMLC and MLC \cite{mlc} to complete a single epoch (in hours) using a single A6000 GPU. 
We find out that MLC \cite{mlc} takes $7.241$ hours whereas EMLC takes $3.67$ hours with $k=1$ and $3.467$ hours with $k=5$.
Our approach is twice as fast as MLC \cite{mlc} despite using a noticeably larger teacher model.
This indicates that our meta-gradient computation is extremely efficient, 
and allows the deployment of our method to large scale datasets.
\section{Discussion}
In this paper we proposed EMLC -- an enhanced meta-label correction framework for the learning with noisy labels problem.
We proposed new meta-gradient approximations for both the single and multi-step optimization strategies and showed that they can be computed very efficiently.
We further offered a teacher architecture that is better aimed to handle label noise.
We present a novel adversarial noise injection mechanism and train the teacher architecture using both regular supervision and meta-supervision.

EMLC surpasses previous methods on benchmark datasets, 
demonstrating an extraordinary performance improvement of $1.52\%$ on the real noise dataset Clothing1M.

Our proposed meta-learning strategies are accurate and fast in terms of computation and speed of convergence.
As such, despite their application to LNL, we persume that our proposed meta learning strategies can be exploited in other distinct meta-learning problems.

\clearpage
{\small
\bibliographystyle{ieee_fullname}
\bibliography{egbib}
}

\newpage
\appendix

\section{Proofs}
\subsection{Proof of Proposition 1} \label{proof:single-step}
\begin{proof}
The student's update rule parameterized by $\alpha$ is given by:
\begin{align}
    \label{eqn:os}
    w^{(t+1)}(\alpha) = w^{(t)} - \eta_w \grad{w} \til{\loss}(w, \alpha) \bigg\rvert_{w=w^{(t)}}
\end{align}

Differentiating \cref{eqn:os} w.r.t $\alpha$ yields:

\begin{align}
    \label{eqn:osmixed}
    \frac{d w^{(t+1)}}{d \alpha} \bigg\rvert_{\alpha=\alpha^{(t)}} = -\eta_w \grad{w \alpha} \til{\loss}(w, \alpha) \bigg\rvert_{(w^{(t)}, \alpha^{(t)})}
\end{align}

We now can compute the meta-gradient explicitly using the chain rule as follows:

\begin{align}
    \frac{d \loss}{d \alpha} \bigg\rvert_{\alpha=\alpha^{(t)}} =
    \frac{d \loss}{d w^{(t+1)}} \bigg\rvert_{w^{(t+1)}=w^{(t+1)}} \cdot
    \frac{d w^{(t+1)}}{d \alpha} \bigg\rvert_{\alpha=\alpha^{(t)}}  
\end{align}

Where $w^{(t+1)}$ is obtained from the updated student.

Substituting \cref{eqn:osmixed} we obtain the desired result.
\end{proof}

\subsection{Proof of Proposition 2} \label{proof:multi-step}
\begin{proof}    
In the case $k>1$, we have: 
\begin{align}
    \label{eqn:recursion_base}
    w^{(t-k+2)}(\alpha) = w^{(t-k+1)} - \eta_w \grad{w} \til{\loss}(w^{(t-k+1)}, \alpha)
\end{align}
Recall that we neglect the dependency of $w^{(t-k+1)}$ on $\alpha$.
In addition, for all $t-k+2 \leq \tau \leq t$:

\begin{align}
    \label{eqn:ms}
    w^{(\tau+1)}(\alpha) = w^{(\tau)}(\alpha) - \eta_w \grad{w} \til{\loss}(w^{(\tau)}(\alpha), \alpha)
\end{align}

Recall also that:
\begin{align}
    \label{eqn:alpha}
    \forall t-k+1 \leq \tau \leq t: \alpha^{(\tau)} = \alpha^{(t)}
\end{align}

We denote:

\begin{align}
    & H_{w \alpha}^{(\tau)} :=  \grad{w \alpha} \til{\loss}(w^{(\tau)}, \alpha^{(\tau)}) \\
    & H_{w w}^{(\tau)} :=  \grad{w w} \til{\loss}(w^{(\tau)}, \alpha^{(\tau)})
\end{align}

deriving \cref{eqn:ms} w.r.t $\alpha$ (considering the derivative at $\alpha=\alpha^{(t)}$), using the chain rule again and substituting \cref{eqn:alpha} yields:

\begin{align}
    \frac{d w^{(\tau+1)}}{d \alpha} =
    \frac{d w^{(\tau)}}{d \alpha} - 
    \eta_w [H_{w w}^{(\tau)} \cdot \frac{d w^{(\tau)}}{d \alpha} + H_{w \alpha}^{(\tau)} ]
\end{align}

We rewrite the above equation as:

\begin{align}
    \frac{d w^{(\tau+1)}}{d \alpha} =
    [I - \eta_w H_{w w}^{(\tau)}] \cdot \frac{d w^{(\tau)}}{d \alpha} -  \eta_w H_{w \alpha}^{(t)}
\end{align}

If we now approximate $H_{w w}^{(\tau)} \approx I$ we get:

\begin{align}
    \frac{d w^{(\tau+1)}}{d \alpha} =
    (1-\eta_w) \frac{d w^{(\tau)}}{d \alpha} -  \eta_w H_{w \alpha}^{(\tau)}
\end{align}

By setting $\gamma_w = 1-\eta_w$,
opening up the recursion formula
and using \cref{eqn:recursion_base} at the end of the recursion,
we get the desired result.

\end{proof}

\subsection{Proof of Proposition 3} \label{proof:jacobians}

\begin{proof}
This follows from the definition of cross-entropy and simple differentiation rules.
Indeed,
\begin{align}
    & \til{\ell}_i(w,\alpha) = 
    CE(q_{\alpha}(y|x_i, \til{y}_i),p_w(y|x_i)) = \\
    & - \sum_{c=1}^C (q_{\alpha}(y=c|x_i, \til{y}_i) \cdot \log(p_w(y=c|x_i)))) = \\
    \label{eqn:inner-prod}
    & - \left \langle q_{\alpha}(y|x_i, \til{y}_i) , \log p_w(y|x_i)) \right \rangle
\end{align}
Now, for two general differential functions 
$f(w): \mathbb{R}^W \rightarrow \mathbb{R}^C$,
and $g(\alpha): \mathbb{R}^A \rightarrow \mathbb{R}^C$
consider $h(w,\alpha) = \left \langle f(w), g(\alpha)\right \rangle$.
Then:
\begin{align}
    \grad{w} h(w, \alpha) = 
    \grad{w} \left \langle f(w), g(\alpha)\right \rangle = 
    (g(\alpha))^T J_w(f)
\end{align}
Differentiating both sides of the equation w.r.t $\alpha$ yields:
\begin{align}
    \grad{w \alpha} h(w, \alpha) = 
    \frac{d \grad{w} h}{d g} \cdot \frac{d g}{d \alpha} = 
    [J_w(f)]^T [J_\alpha(g)]
\end{align}
And hence $\grad{w \alpha} h(w, \alpha)$ exists and equals to $[J_w(f)]^T[J_\alpha(g)]$.
Substituting $f,g$ from \cref{eqn:inner-prod} in the above equation yields the desired result. 
\end{proof}

\section{Comparison of the Meta Gradient Computation} \label{appendix:meta_grad_comp}

We compare the differences of our FPMG algorithm and MLC \cite{mlc} in terms of the quality and efficiency of the meta-gradient computation in \cref{tbl:meta_grad_comp}.
Prominently, FPMG avoids computing second order derivatives which yield a large memory and computation overhead.

\begin{table}
    	\centering
    \setlength\extrarowheight{2pt}
	\begin{tabular}	{c | c  c}
        \toprule
			\textbf{Criterion} &
                \textbf{FPMG} &
                \textbf{MLC} \\
        \midrule
                Exact GD recursion & \checkmark & \xmark \\
                Exact mixed Hessian $H_{w\alpha}$ & \checkmark & \checkmark \\
			Avoids second-order derivative & \checkmark & \xmark \\
                Approximation of $H_{ww}$ & $H_{ww} \approx I$ & $H_{ww} \approx I$ \\
		\bottomrule
	\end{tabular}
	\caption
		{
		Comparison of FPMG and MLC \cite{mlc} regarding the meta-gradient quality and efficiency of the computation.
		}
	\label{tbl:meta_grad_comp}
\end{table}

\section{Additional Experiments}

\subsection{Ablation studies}
\label{sec:ablations}

We perform an ablation study on the different components of our method.
Each setting is validated on the Clothing1M dataset.
The results are summarized in \cref{tbl:ablations}.

As can be observed from \cref{tbl:ablations}, our approach possesses a very strong baseline.
Each of our proposed components incrementally improves the effectiveness, as expected.
Notably, it can be observed that the meta-regularization and the artificial corruption were crucial for the success of our method.

\begin{figure}
\centering
\includegraphics[width=0.5\textwidth]{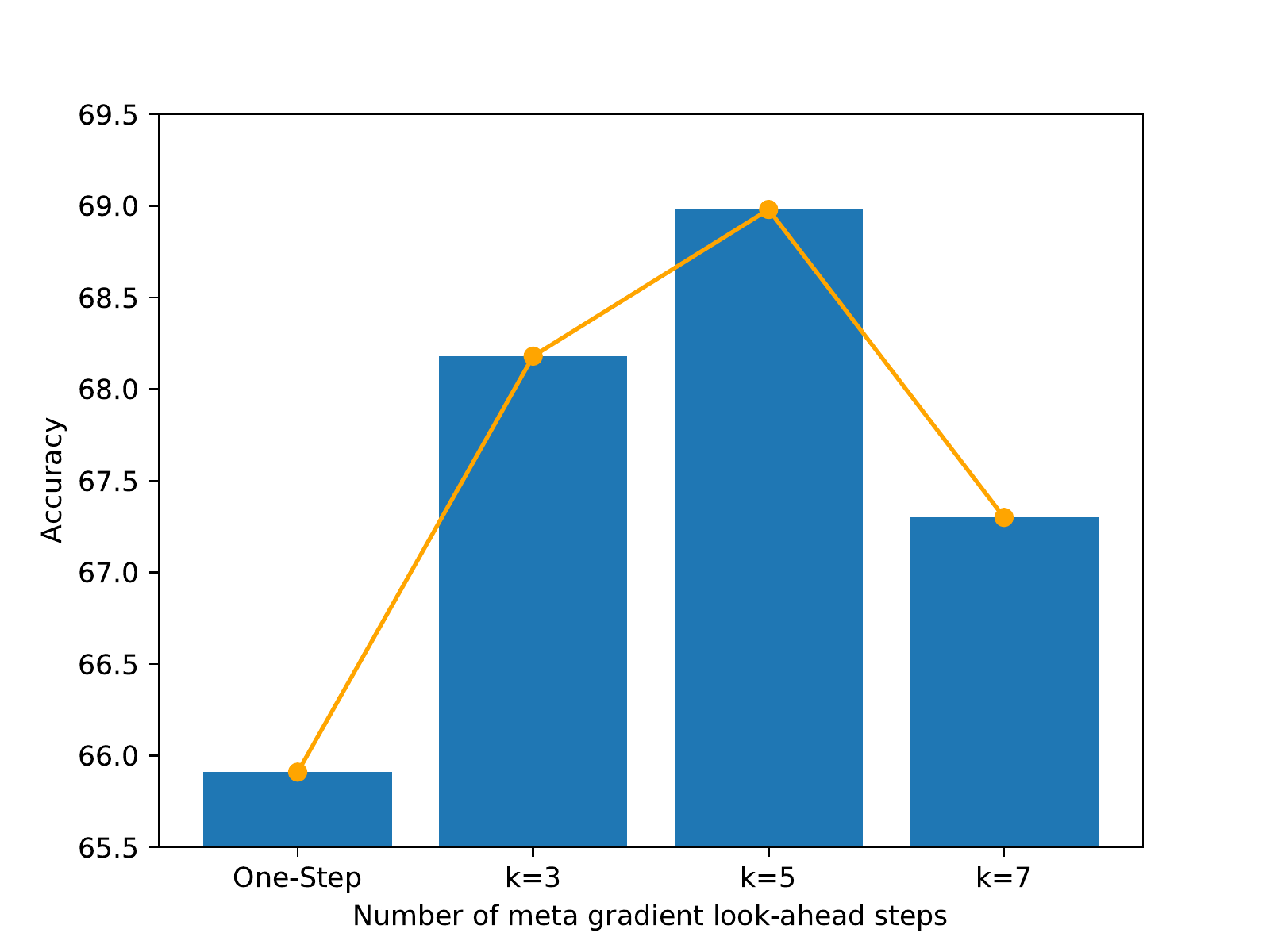}
\caption{Ablation study on the number of look-ahead steps, measuring the effectiveness on the CIFAR-100 dataset with $50\%$ symmetric noise.
%Revise the caption
}
\label{fig:steps}
\end{figure}

\begin{figure}
\centering
\includegraphics[width=0.5\textwidth]{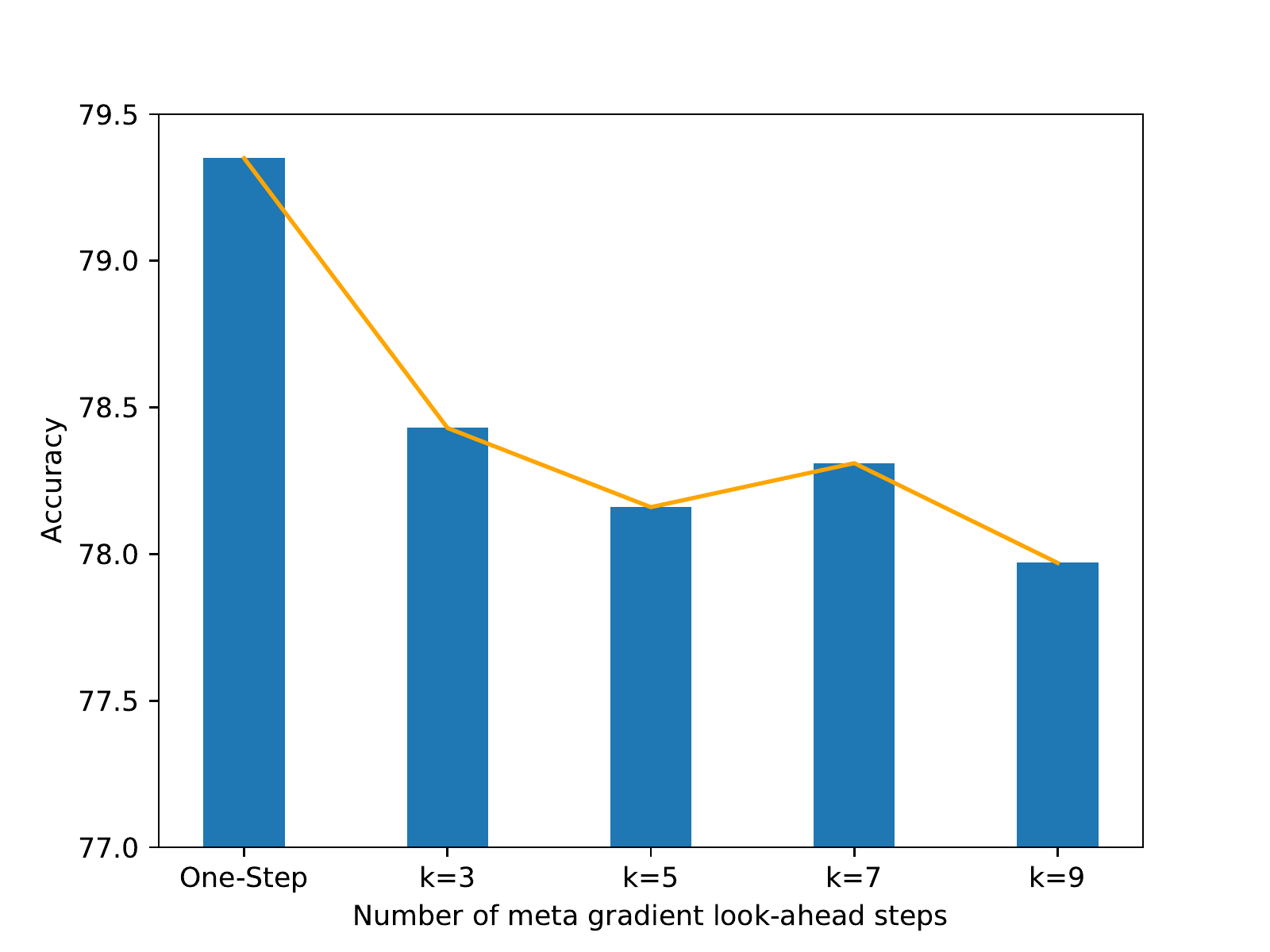}
\caption{Ablation study on the number of look-ahead steps, measuring the effectiveness on the Clothing1M dataset.
%Revise the caption
}
\label{fig:clothing_steps}
\end{figure}

Regarding the number of look-ahead steps, we perform two ablative experiments.
Following the finding that the multi-step strategy outclassed the single-step strategy in some of the CIFAR experiments, we perform an ablation study on the number of steps in the multi-step strategy on the CIFAR-100 dataset with $50\%$ symmetric noise and a fixed seed.
As can be observed from the results in \cref{fig:steps}, $k=5$ arbitrated to be the best.
In addition, due to the importance of the Clothing1M dataset, we perform an additional ablation study to demonstrate the robustness of our method to varying number of look-ahead steps on a real-world dataset, as presented in \cref{fig:clothing_steps}.

\begin{table}
    	\centering
    \setlength\extrarowheight{2pt}
	\begin{tabular}	{c  c  c | c}
        \toprule
			\thead{\textbf{Meta} \\ \textbf{Regularization}} &
            \thead{\textbf{Corruption} \\ \textbf{Strategy}} &
            \thead{\textbf{Strong} \\ \textbf{Augmentations}} &
            \textbf{Accuracy} \\
        \midrule
			\xmark & \xmark & \xmark & 74.35 \\
            \cmark & \xmark & \xmark & 77.52 \\
            \cmark & Rand. & \xmark & 78.51 \\
            \cmark & Adv. & \xmark & 78.60 \\
            \cmark & Adv. & \cmark & \textbf{79.35} \\
            % \cmark & Adv. & \cmark & $k=7$ & 77.02 \\
            % \cmark & Adv. & \cmark & $k=5$ & 78.16 \\
            % \cmark & Adv. & \cmark & $k=3$ & 78.47 \\
            
		\bottomrule
	\end{tabular}
	\caption
		{
		Ablation study on the distinct components of EMLC.
        We validate the effectiveness of each combination by considering test accuracy (\%) on the Clothing1M dataset.
        We verify the effectiveness of the meta-learning regularization,
        the effectiveness of the proposed proactive noise injection 
        and the benefit of applying AutoAugment to the labeled data.
        The corruption strategy values might be either none (\xmark -- in which case, the $BCE$ loss is omitted), random (rand.) or adversarial (adv.). 
		}
	\label{tbl:ablations}
\end{table}

\subsection{Teacher's Label Recovery}
\begin{figure}
\centering
\includegraphics[width=0.5\textwidth]{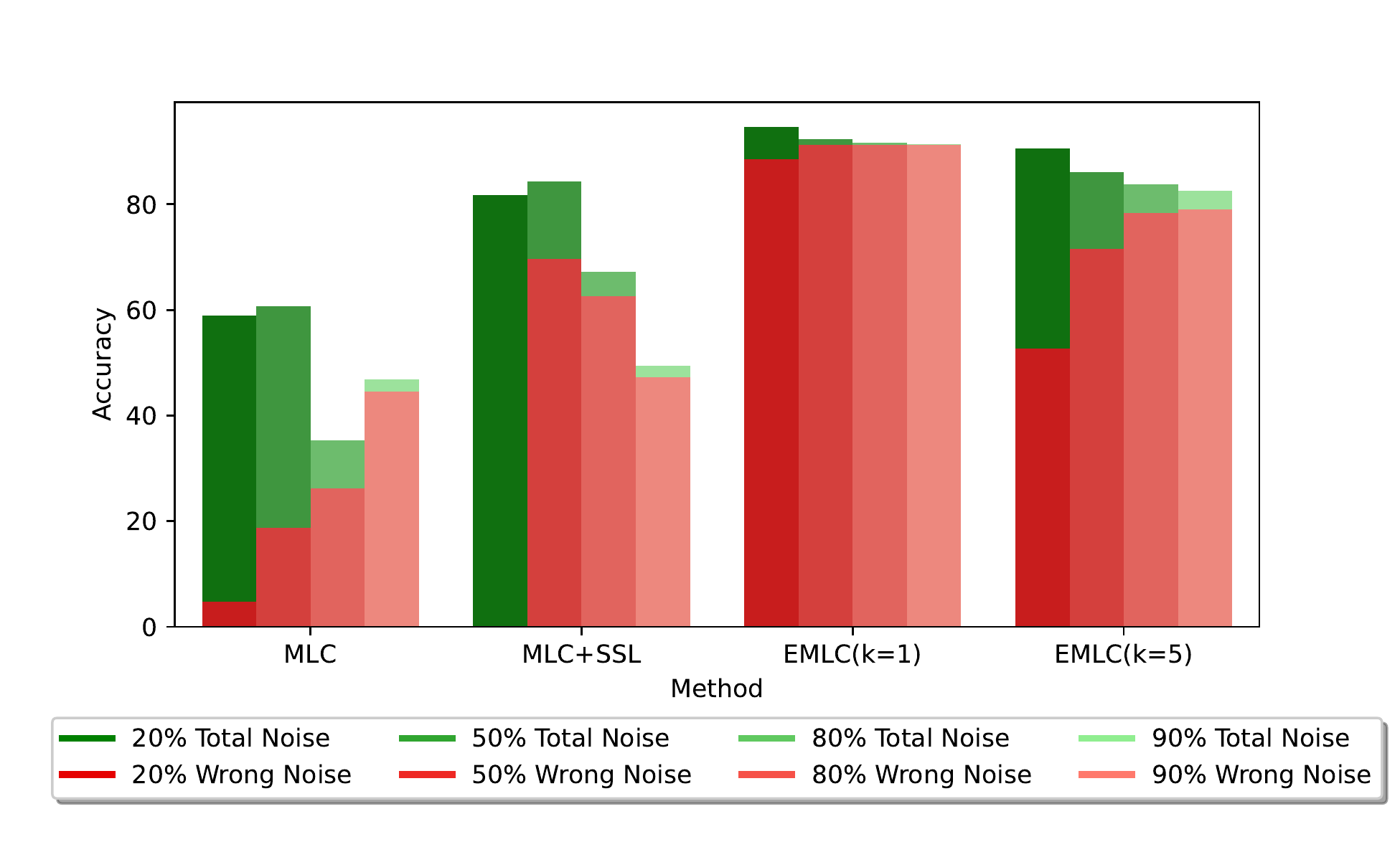}
\caption{Comparison of the teacher’s label recovery rate (total and wrongly annotated samples) on the CIFAR-10 dataset with different noise levels.
%Revise the caption
}
\label{fig:label_recovery}
\end{figure}

To further verify the teacher’s ability to cleanse the training labels, we compare the teacher's label recovery rate (total and wrongly annotated samples) on the CIFAR-10 dataset with different noise levels of EMLC against MLC \cite{mlc} and MLC with self-supervised pretraining in \cref{fig:label_recovery}.

\section{A probabilistic interpretation of the teacher}

The goal of the teacher is to model the conditional distribution of the true label $y$ given the sample $x$ and its noisy label $\Tilde{y}$, namely $p(y|x,\Tilde{y})$.

The above conditional distribution can be decomposed as follows:

\begin{align}
    & p(y|x,\Tilde{y}) = \\
    & w \cdot p(y|x,\Tilde{y}, \mathcal{A}) + (1-w) \cdot p(y|x,\Tilde{y}, \mathcal{A}^c) = \\
    & w \cdot \delta_{\Tilde{y} y} + (1-w) \cdot p(y|x,\Tilde{y}, \mathcal{A}^c)
\end{align}

where $\mathcal{A}$ is the event of $y = \Tilde{y}$ and $w := p(\mathcal{A} | x,\Tilde{y})$.

In our teacher architecture, we model $w$ directly. 
However, we approximate $p(y|x,\Tilde{y}, \mathcal{A}^c) \approx p(y|x, \mathcal{A}^c) \approx p(y|x)$ and model $p(y|x)$ instead.
In the first approximation, we throw away the conditioning on $\Tilde{y}$,  intuitively ignoring the cases in which $y|x$ is dependent of $\tilde{y}$ whenever the label is corrupted.
While the second approximation is technically unnecessary (as $p(y|x, \mathcal{A}^c)$ can be easily modeled by masking the \textit{SoftMax} layer),
we found out that the latter modeling was better in practice since it effectively enhances the weight of hard clean samples.

\section{Further Details of the Experimental Setting}

\subsection{Symmetric and Asymmetric Artificial Noise} \label{appendix:sym_vs_asym}

In the \textit{Symmetric noise} setting, a fraction of $\epsilon$ samples are chosen randomly.
The samples are then assigned with a random label selected uniformly over the classes.
The expected fraction of \textit{wrong samples} is effectively smaller than $\epsilon$.
This noise definition is convenient, as $\epsilon = 1$ would mean that the label assignment is entirely random.
In the \textit{Asymmetric noise} setting proposed by Partini \etal \cite{partini}, a fraction of $\epsilon$ samples are chosen randomly.
Then, their label is replaced by a label of a visually similar category,
to better model real-world noise.
In the CIFAR-10 dataset, this is done by applying the following transitions:
TRUCK $\rightarrow$ AUTOMOBILE, BIRD $\rightarrow$ AIRPLANE, DEER $\rightarrow$ HORSE, CAT $\leftrightarrow$ DOG.
In the CIFAR-100 dataset, each label is changed to its successor(circularly) in its super-class.

\subsection{Implementation Details and Hyperparameters} \label{appendix:implementation}
Further experimental details are presented in \cref{tbl:experimental}.

\begin{table*}
    	\centering
    \setlength\extrarowheight{2pt}
	\begin{tabular}	{l | c  c c}
        \toprule
			\textbf{Dataset} & 
            \textbf{CIFAR-10} & 
            \textbf{CIFAR-100} & 
            \textbf{Clothing1M}\\
        \midrule
			\textbf{Architecture} & ResNet-34 & ResNet-50 & ResNet-50 \\
            \textbf{Pretraining} & SimCLR & SimCLR & ImageNet \\
            \textbf{MLP hidden layers} & 1 & 1 & 1 \\
            \textbf{MLP hidden units} & 128 & 128 & 128 \\
            \textbf{Noisy batch size} & 128 & 128 & 1024 \\
            \textbf{Clean batch size} & 32 & 32 & 1024 \\
            \textbf{Epochs} & 15 & 15 & 3 \\
            \textbf{Noisy augmentations} &
            \thead{Horizontal Flip \\ Random Crop} &
            \thead{Horizontal Flip \\ Random Crop} & None \\
            
            \textbf{Clean augmentations} &
            \thead{Horizontal Flip \\ Random Crop \\ CIFAR AutoAugment} & \thead{Horizontal Flip \\ Random Crop \\ CIFAR AutoAugment} & \thead{Horizontal Flip \\ ImageNet AutoAugment} \\
            
            \textbf{Optimizer} & SGD & SGD & SGD \\
            \textbf{Scheduler} & None & None & LR-step \\
            \textbf{Momentum} & 0.9 & 0.9 & 0.9 \\
            \textbf{Weight decay} & 0 & 0 & 0 \\
            \textbf{Initial LR} & 0.02 & 0.02 & 0.1 \\
            \textbf{Number of GPUs} & 1 & 1 & 8 \\
            
		\bottomrule
	\end{tabular}
	\caption
		{
		Experimental setup and hyper-parameters.
		}
	\label{tbl:experimental}
\end{table*}

\end{document}